\documentclass[letterpaper]{article} 
\usepackage{aaai23}  
\usepackage{times}  
\usepackage{helvet}  
\usepackage{courier}  
\usepackage[hyphens]{url}  
\usepackage{graphicx} 
\usepackage{natbib}  
\usepackage{caption} 
\usepackage{algorithm}
\usepackage[noend]{algpseudocode}

\usepackage{newfloat}
\usepackage{listings}
\DeclareCaptionStyle{ruled}{labelfont=normalfont,labelsep=colon,strut=off} 
\floatstyle{ruled}
\newfloat{listing}{tb}{lst}{}
\floatname{listing}{Listing}
\usepackage{dsfont}
\usepackage{amsmath}
\usepackage{amssymb}
\usepackage[english]{babel}
\usepackage{amsthm}
\usepackage{multirow}
\usepackage{makecell}
\usepackage{graphicx}
\usepackage{verbatim}
\usepackage{booktabs}
\usepackage{arydshln}
\usepackage{subcaption}
\usepackage{xcolor}
\usepackage{paralist}
\usepackage{colortbl}
\usepackage{cleveref}
\theoremstyle{definition}
\newtheorem{definition}{Definition}
\theoremstyle{remark}

\newcommand{\colorb}[1]{\textcolor[RGB]{10,30,91}{#1}}
\newcommand{\colorr}[1]{\textcolor[RGB]{165,42,42}{#1}}
\newcolumntype{C}[1]{>{\centering\let\newline\\\arraybackslash\hspace{0pt}}m{#1}}

\title{Improving Simultaneous Machine Translation with Monolingual Data}
\author{
    Hexuan Deng\textsuperscript{\rm 1}\thanks{Work was done when Hexuan was interning at JD
Explore Academy.},
    Liang Ding\textsuperscript{\rm 2},
    Xuebo Liu\textsuperscript{\rm 1}\thanks{Corresponding author: Xuebo Liu.},
    Meishan Zhang\textsuperscript{\rm 1},
    Dacheng Tao\textsuperscript{\rm 2},
    Min Zhang\textsuperscript{\rm 1}
}
\affiliations{
    \textsuperscript{\rm 1}Institute of Computing and Intelligence, Harbin Institute of Technology, Shenzhen, China
    \\
    \texttt{22s051030@stu.hit.edu.cn, \{liuxuebo,zhangmeishan,zhangmin2021\}@hit.edu.cn} 

    \textsuperscript{\rm 2}JD Explore Academy, JD.com Inc. 
    \\
    \texttt{dingliang1@jd.com, dacheng.tao@gmail.com}
}

\begin{document}

\maketitle

\begin{abstract}
Simultaneous machine translation (SiMT) is usually done via sequence-level knowledge distillation (Seq-KD) from a full-sentence neural machine translation (NMT) model. 
However, there is still a significant performance gap between NMT and SiMT. 
In this work, we propose to leverage monolingual data to improve SiMT, which trains a SiMT student on the combination of bilingual data and external monolingual data distilled by Seq-KD. 
Preliminary experiments on En$\Rightarrow$Zh and En$\Rightarrow$Ja news domain corpora demonstrate that monolingual data can significantly improve translation quality (e.g., +3.15 BLEU on En$\Rightarrow$Zh).
Inspired by the behavior of human simultaneous interpreters, we propose a novel monolingual sampling strategy for SiMT, considering both chunk length and monotonicity. 
Experimental results show that our sampling strategy consistently outperforms the random sampling strategy (and other conventional typical NMT monolingual sampling strategies) by avoiding the key problem of SiMT -- hallucination, and has better scalability. 
We achieve +0.72 BLEU improvements on average against random sampling on En$\Rightarrow$Zh and En$\Rightarrow$Ja. 
Data and codes can be found at~\url{https://github.com/hexuandeng/Mono4SiMT}.
\end{abstract}

\section{Introduction}

Simultaneous machine translation (SiMT) \citep{LearningTranslateRealtime_2017,STACLSimultaneousTranslation_2019,MonotonicInfiniteLookback_2019,SimultaneousTranslationPolicies_2020} has been proposed to generate real-time translation by starting decoding before the source sentence ends. 
However, generation conditioned on the partial source sentence prevents a model from properly capturing the whole semantics, especially for distant languages, e.g., English and Japanese \cite{SyntaxbasedRewritingSimultaneous_2015,ImprovingSimultaneousTranslation_2021}.
In response to this problem, motivated by the recent success of non-autoregressive translation, sequence-level knowledge distillation (Seq-KD,~\citealp{SequenceLevelKnowledgeDistillation_2016}) becomes the preliminary step for training SiMT models, with a full-sentence neural machine translation (NMT) model as the teacher~\cite{SimulSpeechEndtoEndSimultaneous_2020,FutureGuidedIncrementalTransformer_2021}, which helps to generate monotonous knowledge by reducing data complexity~\cite{UnderstandingKnowledgeDistillation_2020}.

Although Seq-KD narrows the gap between full-sentence NMT teachers and SiMT students, the performance gap is still significant.
Techniques like self-training~\cite{ExploitingSourcesideMonolingual_2016,SelfTrainingSamplingMonolingual_2021a} are known to effectively improve machine translation performance by using large-scale monolingual data.
However, to the best of our knowledge, improving SiMT through semi-supervised learning has not been well validated yet.

To this aim, we leverage the monolingual data to perform Seq-KD and train the SiMT student model on the combination of distilled monolingual and bilingual data. 
Exploiting monolingual data for SiMT provides appealing benefits. First, the monolingual data
and bilingual data in machine translation are generally complementary to each other~\cite{ImprovingNeuralMachine_2016,ExploitingSourcesideMonolingual_2016,ImprovingNonautoregressiveNeural_2020a,ding2022redistributing}. Accordingly, using monolingual for SiMT transfers both the knowledge of the bilingual data (implicitly encoded in the full-sentence NMT teacher) and that of monolingual data, maintaining the merit of Seq-KD to reduce the complexity of the bilingual data. Secondly, the amount of available monolingual data is several orders of magnitude larger than that of bilingual data, offering great potential to enjoy attractive expandability.

However, unlike NMT, it is difficult for SiMT to handle long-distance reordering~\cite{ImprovingNonautoregressiveNeural_2020a}. Therefore, the pseudo-targets generated by the full-sentence NMT teacher model are not always suitable for SiMT. 
Inspired by strategies used in human simultaneous interpretation, e.g., finer segments and monotonic alignments~\cite{InterpreteseVsTranslationese_2016}, we propose novel strategies for sampling monolingual data suitable for SiMT, considering both the chunk lengths and monotonicity.
We validate our strategy on several large-scale datasets of news domain (En$\Rightarrow$Zh and En$\Rightarrow$Ja). 
Our contributions are as follows:

\begin{compactitem}
\item We empirically demonstrate that using monolingual data is beneficial to SiMT systems.
\item Our monolingual data sampling strategy for SiMT significantly outperforms the random sampling and conventional NMT monolingual sampling strategies, especially evaluating at low latency.
\item Our strategy effectively alleviates the key issue of SiMT, i.e., hallucination problem, and has high expandability, e.g., enlarging the scale of monolingual data consistently improves performance.
\end{compactitem}

The paper is an early step in exploring monolingual data for SiMT, which can narrow the performance gap between SiMT models and the SOTA full-sentence NMT models. We hope the promising effect of the monolingual sampling strategy on SiMT can encourage further investigation and pave the way toward more effective SiMT models.

\section{Background and Related Work}

\paragraph{Simultaneous Machine Translation}
Full-sentence NMT models use Seq2seq framework, where the encoder takes the source sentence $\boldsymbol{x}=(x_1,...,x_m)$ as input, and outputs hidden state $\boldsymbol{h}=(h_1,...,h_m)$. Then, the decoder iteratively predicts the next token $y_t$ based on the hidden state and previously generated tokens until the end of the sequence:

\begin{equation}
\hat{y_t}=\underset{y_t}{\textrm{argmax}}\ p\left(y_t \mid \boldsymbol{x}, \boldsymbol{y}_{<t} ; \boldsymbol{\theta}\right)
\end{equation}

In SiMT, we cannot access the entire source sentence when decoding. \citet{STACLSimultaneousTranslation_2019} propose a simple but efficient wait-$k$ policy to balance translation quality and delay. Specifically, it first reads $k$ words, then loops to read and write a word until the end of the sentence:

\begin{equation}
\hat{y_t}=\underset{y_t}{\textrm{argmax}}\ p\left(y_t \mid \boldsymbol{x}_{\leq g_{\text{wait-}k}(t)}, \boldsymbol{y}_{<t} ; \boldsymbol{\theta}\right)
\end{equation}

\noindent where $g_{\text{wait-}k}(t)=\min \{k+t-1,|\boldsymbol{x}|\}$ indicates the number of source words that can be seen when predicting word $y_t$ under the wait-$k$ policy.

Several works have been proposed to narrow the gap between SiMT and NMT datasets. \citet{SyntaxbasedRewritingSimultaneous_2015} use handwriting language-specific rules based on syntax trees to generate pseudo-targets with fewer reordering, but it requires linguistic knowledge and is difficult to transfer to other language pairs. \citet{zhang-etal-2020-learning-adaptive} use the sentence-aligned parallel corpus to train an NMT model and generate pseudo-targets with a policy according to the attention of NMT, while \citet{ImprovingSimultaneousTranslation_2021} directly use the test-time wait-$k$ policy, which significantly reduces the anticipation rate while simplifies the computational complexity. \citet{MonotonicSimultaneousTranslation_2021a} employ a method based on chunk-wise reordering and NAT refinement to generate monotonic and smooth references.
Unlike the above approaches that utilize bilingual data effectively, our study is the first work to investigate how to improve SiMT with large-scale monolingual data, which is orthogonal to the above approaches.

\paragraph{Semi-Supervised NMT}
NMT models are data hungry, and the translation quality highly depends on the quality and quantity of parallel corpus~\cite{SixChallengesNeural_2017a,liu2020norm}. Researchers thus turn to investigate the effects of using large-scale monolingual data~\cite{ExploitingSourcesideMonolingual_2016,UsingTargetsideMonolingual_2017,edunov2018understanding,ding2021usyd} with semi-supervised learning~\cite{zhu2009introduction}. The general process follows several steps: 1) train a base model with bilingual data; 2) decode the large-scale monolingual data with the pre-trained base model to obtain the synthetic data; and 3) retrain the model with the concatenation of bilingual and synthetic data.
Designing an effective monolingual sampling strategy is at the core of the process.
\citet{IntelligentSelectionLanguage_2010a} select in-domain monolingual samples through the source language model. \citet{BackTranslationSamplingTargeting_2018a} improve the prediction accuracy of the model by selecting sentences with lower frequency words, while \citet{SelfTrainingSamplingMonolingual_2021a} achieve a similar purpose by sampling monolingual data with high uncertainty.
While semi-supervised learning shows great success in full-sentence translation, few works explore the effects of using monolingual data for SiMT. We take the first step to investigate SiMT-aware monolingual sampling strategies and their best combination and provide a comprehensive discussion to show the scalability of our approach.

\section{Monolingual Data Sampling Strategies}

We introduce the sampling strategies and the corresponding metrics, where monolingual data with lower scores are considered more efficient and used for training. The tendency to choose longer sentences is added to all these strategies and will be introduced first.

\paragraph{Sentence Length}
Longer sentences usually contain more information, encouraging the model to make use of more context information \cite{CompetencebasedCurriculumLearning_2019a}. Besides, training with longer sentences can suppress the generation of end signal ``$<$EOS$>$'' and nicely alleviate the early-stop phenomenon in SiMT, where the generating ends are given the incomplete source input. Therefore, in all subsequent sampling strategies, we add long sentence tendency factor $\alpha$ by replacing the sentence length term (or similar item) $|\boldsymbol{x}|$ with $|\boldsymbol{x}|^\alpha$ (or $|\boldsymbol{x}|^{1/\alpha}$), aiming at tending to choose longer sentences while maintaining the effectiveness of the strategy. In our experiments, we set $\alpha=0.5$ as default.

\subsection{Sample Corpora More Suitable for SiMT}

In response to different word order between language pairs, \citet{InterpreteseVsTranslationese_2016} point out that human interpretation often: 1) breaks source sentences into multiple smaller chunks and uses conjunctions for fluently connecting; 2) uses passivization to wait for the source to give the verb without stopping the translation process, especially when from head-final languages (e.g., Japanese) to head-initial languages (e.g., English). Both of them greatly alleviate the problems above while ensuring fluency.

\subsubsection{Chunk Length-Based Strategy}

Inspired by the first phenomenon, the easiest way is to select data with shorter chunks for training to develop its tendencies, aiming at obtaining the same benefits as above. As for chunk extraction, we want to evaluate the chunk length of the current monolingual corpora at the lowest cost rather than extracting meaningful units. Under such consideration, we propose the following two metrics to give a relatively accurate evaluation at a lower cost, which is detailly described in the appendix.

\begin{figure}[t]
\centering
\includegraphics[width=0.4\textwidth]{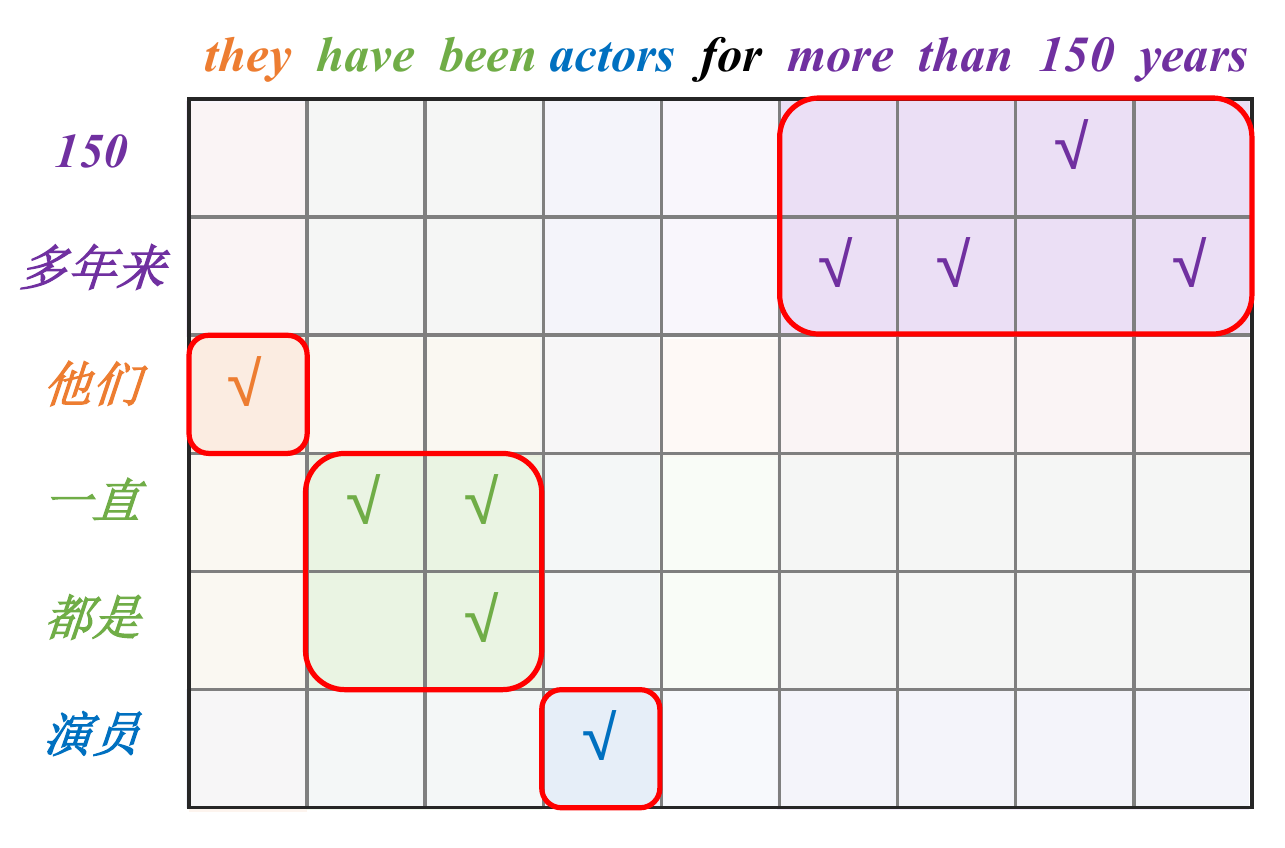}
\caption{Example of alignment-based chunk extraction, where ``$\checkmark$'' means that the source- and target-side tokens are aligned, and the red rectangles are the extracted chunk pairs.}
\label{fig:align}
\end{figure}

Inspired by \citet{HierarchicalPhraseBasedTranslation_2007}, \textit{Alignment-based approach} selects the shortest contiguously aligned block as a chunk, which satisfies that tokens in the source part are aligned with and only with corresponding tokens in the target part and vice versa, while the source part and the target part are contiguous and inseparable. As shown in Figure~\ref{fig:align}, the parts enclosed by the red box are chunks we identified. This method can extract meaningful chunks in most cases but need pseudo-targets and alignments for monolingual data, which is time-consuming.

To extract chunks efficiently, inspired by \citet{UniversityEdinburghSubmission_2021}, we employ source-side language model (LM): \textit{LM-based approach} keeps track of the LM score of the prefix of source sentences and adds token once at a time. If the new LM score is lower than the previous one, the previous prefix will be considered as a chunk. Afterwards, the next word is regarded as the beginning of the sentence, and recursively perform the above steps until the end of the sentence. Although there is no information about pseudo-targets, it can also play a similar or even better effect than the previous method in our experiments (See Table~\ref{tab:sampling_strategies}).

In the calculation of metrics, the numerator is the number of alignments in the source sentence for the alignment-based approach and sentence length for the LM-based approach. We add index $\alpha$ to those numerators as exponents to reflect the long sentence tendency. In this way, for the alignment-based approach, sentences with denser alignments are also tended to be chosen, which intuitively have lower error rates and contain more information, which should also be encouraged. Formally, if we define the total number of chunks in the sentence as $c$ and the numerator as $\ell$, the chunk length-based metric for the sentence is:

\begin{equation}
S_{chunk}=\frac{\ell^\alpha}{c}
\end{equation}

\subsubsection{Monotonicity-Based Strategy}

Inspired by the second phenomenon, we take a straightforward solution to choose sentences with more monotonous alignments directly. Refer to \citet{ImprovingSimultaneousTranslation_2021}, we use $k$-Anticipation Rate ($k$-AR) as metric for monotonicity. Specifically, for each aligned target word $y_j$, it is considered a $k$-anticipation if it is aligned to a source word $x_i$ that is $k$ words behind. The $k$-AR is then calculated as the percentage of $k$-anticipation among all aligned word pairs. Specifically, if the set $\mathcal{A}=\{(i_t,j_t)\}_{t=1}^N$ represents all aligned token-pairs $\boldsymbol{x}_{i_k} \sim \boldsymbol{y}_{j_k}$, the monotonicity-based metric for the sentence is:

\begin{equation}
S_{mono}=\frac{1}{|\mathcal{A}|^{1/\alpha}} \sum_{t=1} ^{|\mathcal{A}|} \mathds{1}[i_t \leq j_t + k]
\end{equation}

\noindent where $\alpha$ is the long sentence tendency factor, which also adds bias for sentences with denser alignments as with the alignment-based approach.

\subsection{Sentence Difficulty}

In traditional NMT, there are some solutions for sampling monolingual data according to difficulty. We choose two of them and add the same long sentence tendency factor $\alpha$ for comparison.

\citet{BackTranslationSamplingTargeting_2018a} propose that monolingual data containing low-frequency words are more conducive to model training. Then \citet{CompetencebasedCurriculumLearning_2019a} use the source-side unigram language model to reflect the tendency to select sentences that are longer and contain more low-frequency words at the same time. In our setup, for monolingual sentence $\boldsymbol{x}=(x_1,...,x_m)$, and the probability $\hat{p}\left(x_i\right)$ of each word $x_i$ occurred in the bilingual corpora, taking into account the tendency to choose long sentences, the frequency metric for the sentence is:

\begin{equation}
S_{\text {rarity }} =-\frac{1}{|\boldsymbol{x}|^\alpha}\sum_{i=1}^{|\boldsymbol{x}|} \log \hat{p}\left(x_i\right)
\end{equation}

\citet{SelfTrainingSamplingMonolingual_2021a} propose a metric based on uncertainty. It first evaluates word level entropy $E$ by using the alignment $\mathcal{A}$ on bilingual corpora to capture the translation modalities of each source token. Specifically, for a given monolingual sentence $\boldsymbol{x}=(x_1,...,x_m)$, if $\mathcal{A}(x_i)$ records all possible target tokens $y_j$ aligned with source token $x_i$, and calculate the translation probability $p(y_j|x_i)$ according to it, the word level entropy is:

$$
E\left(y \mid \mathcal{A}, x_{i}\right)=-\sum_{y_{j} \in \mathcal{A}\left(x_{i}\right)} p\left(y_{j} \mid x_{i}\right) \log p\left(y_{j} \mid x_{i}\right)
$$

For the monolingual data, taking into account the tendency to choose long sentences, its uncertainty metric is:

\begin{equation}
S_{\text{uncer}}=\frac{1}{|\boldsymbol{x}|^\alpha} \sum_{i=1}^{|\boldsymbol{x}|} E\left(y \mid \mathcal{A}, x=x_{i}\right)
\end{equation}

\begin{table}[t]
\centering
\begin{tabular}{cccc}
\toprule
& \bf Raw & \bf KD & \bf KD+Mono.\\
\midrule
&\multicolumn{3}{c}{Teacher: 48.55} \\ \cmidrule{2-4}
\textit{wait-1} & 28.62 & 29.93 & \textbf{35.64} \\
\textit{wait-3} & 35.39 & 36.15 & \textbf{39.82} \\
\textit{wait-5} & 39.07 & 41.14 & \textbf{43.46} \\
\textit{wait-7} & 42.52 & 43.76 & \textbf{45.95} \\
\textit{wait-9} & 44.02 & 45.66 & \textbf{47.51} \\
\textit{Avg.} & 
\makecell{\underline{37.92} \\[-1.5pt] \small{(- / -)}} &
\makecell{\underline{39.33} \\[-1.5pt] \bf \small{(\colorr{+1.41}/\colorb{-})}} &
\makecell{\underline{\textbf{42.48}} \\[-1.5pt] \bf \small{(\colorr{+4.56}/\colorb{+3.15})}} \\[-0.5pt] \bottomrule
\end{tabular}
\caption{\label{tab:distillation_and_mono}
The effects of using monolingual data. ``Raw/ KD'' means the results of original/distilled parallel data, and ``+Mono.'' represents enhancing the model with synthetic data generated by randomly sampled monolingual data. Gains against ``Raw'' and ``KD'' are marked with \colorr{red} and \colorb{blue}, respectively. Average scores on all delays are \underline{underlined}. The best results are \textbf{bold}.}
\end{table}

\section{Experiments}

\subsection{Experimental Setup}

\paragraph{Bilingual Data}
We conduct experiments on two widely-used SiMT language directions: English-Chinese (En$\Rightarrow$Zh) and English-Japanese (En$\Rightarrow$Ja).
To make the experiments convincing, we select resource-rich datasets of news domain:
For En$\Rightarrow$Zh, we use CWMT Corpus\footnote{\url{http://nlp.nju.edu.cn/cwmt-wmt/}}~\cite{chen2019machine} as training data, {NJU-newsdev2018} as the validation set and report results on {CWMT2008, CWMT2009, and CWMT2011}; 
For En$\Rightarrow$Ja, we use {JParaCrawl}\footnote{\url{https://www.kecl.ntt.co.jp/icl/lirg/jparacrawl/}}~\cite{JParaCrawlLargeScale_2020a} and {WikiMatrix}\footnote{\url{https://opus.nlpl.eu/WikiMatrix.php}}~\cite{WikiMatrixMining135M_2021} as training data, {newsdev2020} as the validation set and report results on {newstest2020}. 
Considering the corpora are noisy, we apply a series of filtration rules to them, including 1) empty and duplicated lines, 2) sentence pairs with invalid characters, 3) sentence pairs with too many or too few words, and 4) those with too large bilingual length ratios, etc. After data cleaning, we randomly select a subset of 7$M$ sentence pairs as training data for both En$\Rightarrow$Zh and En$\Rightarrow$Ja. 
We use {SentencePiece}~\cite{SentencePieceSimpleLanguage_2018} to split the training data into subword units~\cite{NeuralMachineTranslation_2016} with 32K merge operations. We publicly release our processed datasets\footnote{\url{https://drive.google.com/drive/folders/1HbzxBD0klgX-EugVGB36CFVdObJJ5Uk7?usp=sharing}}.

\paragraph{Monolingual Data}
We closely follow previous works to randomly select monolingual data from publicly available News Crawl corpus\footnote{\url{http://data.statmt.org/news-crawl}}~\cite{ExploitingSourcesideMonolingual_2016,ExploitingMonolingualData_2019a}. For a fair comparison, the monolingual data used in the main experiments have the same size as the corresponding bilingual data, i.e., 7$M$. To comprehensively investigate the effects of different monolingual sampling strategies in Table~\ref{tab:sampling_strategies}, we randomly sample up to 42$M$ English data from News Crawl 2016 and 2017 in the main experiments. For the at-scale experiments in Table~\ref{tab:scale-mono}, we randomly sample up to 540$M$ sentences from News Crawl 2007$\sim$2017 and News Discussions 2014$\sim$2017. 

\paragraph{Model Training}
We closely follow previous SiMT works~\cite{SimulSpeechEndtoEndSimultaneous_2020,FutureGuidedIncrementalTransformer_2021,NAISTEnglishtoJapaneseSimultaneous_2021a,USTCNELSLIPSystemsSimultaneous_2021a,VolctransNeuralSpeech_2021a} to adopt sequence-level knowledge distillation~\cite{SequenceLevelKnowledgeDistillation_2016} for all systems.
Specifically, we train a full-sentence \textsc{Base} Transformer~\cite{AttentionAllYou_2017} as the teacher on the original bilingual dataset, then perform beam-search decoding for the source side of the original bilingual data or newly introduced monolingual data to generate the distilled data.
The student SiMT model follows the \textsc{Base} model, except for using causal encoders and wait-$k$ policy. To investigate the effects of a better teacher, we use full-sentence \textsc{Big} Transformer at Table~\ref{better-teacher}. Note that we train all models with identical training steps.

We use the \textit{SacreBLEU}~\cite{CallClarityReporting_2018} to measure the translation quality and \textit{SimulEval} \cite{SIMULEVALEvaluationToolkit_2020} to measure the latency for each delay under the \textit{wait-$k$}~\cite{STACLSimultaneousTranslation_2019} policy, and also report the averaged BLEU for different delays to avoid stochasticity.
The CWMT test sets have up to 3 references. Thus, we report the 3-reference BLEU score. We use \textit{fast-align}~\cite{SimpleFastEffective_2013} to extract the alignment information for sentences in Table~\ref{tab:anapolicy}, and strategies more suitable for SiMT, and use \textit{KenLM}~\cite{ScalableModifiedKneserNey_2013} to calculate source language model score in chunk length-based strategy.

\begin{figure}[t]
\centering
\includegraphics[width=0.46\textwidth]{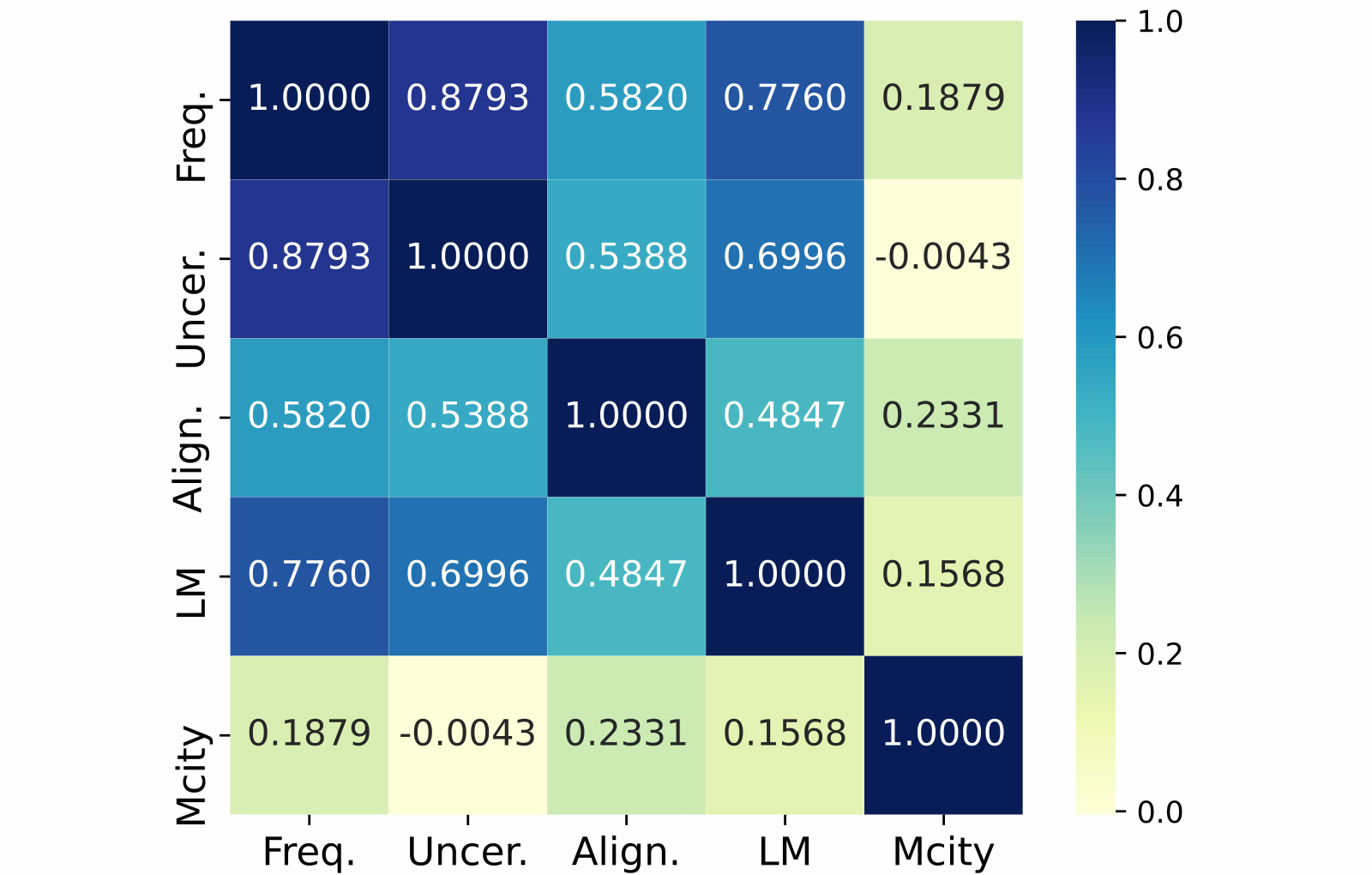}
\caption{
Covariance matrix between different sampling strategies. We score the monolingual dataset using different strategy metrics and calculate the correlation coefficient between different scores. ``Freq.'' and ``Uncer.'' are sentence difficulty metrics estimated with word frequency and uncertainty, respectively. ``Align.'' and ``LM'' are chunk length metrics using alignment-based and LM-based approaches, respectively. ``Mcity'' is monotonicity metric using 3-anticipation rate. The same notations are used in following-up tables.}
\label{fig:corref}
\end{figure}

\begin{table*}[t]
\centering
\begin{tabular}{lccccccc}
\toprule
\bf Strategy & \bf \textit{wait-1} & \bf\textit{wait-3} & \bf\textit{wait-5} & \bf\textit{wait-7} & \bf\textit{wait-9} & \bf\textit{Avg.} & \textbf{$\mathbf{\Delta}$} \\ \midrule
Random & 35.64 & 39.82 & 43.46 & 45.95 & 47.51 & 42.48 \\ \midrule
Frequency-Based Sentence Difficulty Strategy & \bf 36.69 & 40.78 & 44.11 & 46.12 & 47.76 & 43.09 & +0.61 \\
Uncertainty-Based Sentence Difficulty Strategy & 36.26 & 40.95 & 43.33 & 46.30 & 47.57 & 42.88 & +0.40 \\ \midrule
Alignment-Based Chunk Length Strategy & 36.62 & 41.20 & 43.68 & \bf 46.85 & \bf 48.05 & \bf 43.28 & \bf +0.80 \\
LM-Based Chunk Length Strategy & 36.37 & \bf 41.70 & \bf 44.12 & 45.92 & 47.94 & 43.21 & +0.73 \\
Monotonicity-Based Strategy & 35.97 & 40.25 & 42.88 & 45.65 & 46.80 & 42.31 & -0.17 \\ \bottomrule
\end{tabular}
\caption{\label{tab:sampling_strategies}
The effect of different sampling strategies. Since our proposed strategy and baseline belong to the same policy, there is almost no difference in latency. Therefore, we display the results in the form of table to highlight the details of the improvement in translation quality. Improvements against random sampling ``Random'' are in column \textbf{$\mathbf{\Delta}$}.}
\end{table*}

\subsection{Empirical Findings}

In this section, we comprehensively conduct preliminary studies on CWMT En$\Rightarrow$Zh to show 1) the necessity of using monolingual data, 2) the superiority of our proposed SiMT-aware monolingual sampling strategies, and 3) the best strategy combination as our default method.

\paragraph{Monolingual data significantly improves SiMT.}
In order to explore the effect of adding monolingual data, we add the synthetic data generated by randomly sampled monolingual sentences to the distilled parallel data with a ratio of 1:1.
We report the results of original parallel data (``Raw'') for reference.
As shown in Table~\ref{tab:distillation_and_mono}, we can see that distillation improves the SiMT with +1.41 BLEU points on average, and leveraging the randomly sampled monolingual data further pushes the BLEU points by a large margin, i.e., +3.15, especially for the low-latency settings, e.g., +5.71 for wait-1. This confirms the effectiveness of monolingual data for SiMT and urges us to investigate better sampling strategies for monolingual data.

\begin{table}[t]
\centering
\begin{tabular}{cC{1.3cm}cC{1.3cm}c}
\toprule
 & \bf Chunk (Align.) & \bf +Mcity & \bf Chunk (LM) & \bf +Mcity \\ \midrule
\textit{wait-1} & \makecell{36.62 \\[-2.5pt] \small{(-)}} & 
\makecell{\textbf{37.08} \\[-2.5pt] \bf \small{(\colorr{+0.46})}}
& \makecell{36.37 \\[-2.5pt] \small{(-)}} & 
\makecell{\textbf{37.40} \\[-2.5pt] \bf \small{(\colorr{+1.03})}}
\\
\textit{wait-3} & \bf 41.20 & 41.10 & \bf 41.70 & 40.49 \\
\textit{wait-5} & 43.68 & \textbf{44.28} & 44.12 & \textbf{44.44} \\
\textit{wait-7} & \bf 46.85 & 46.46 & 45.92 & \textbf{46.27} \\
\textit{wait-9} & \bf 48.05 & 47.69 & 47.94 & \textbf{48.00} \\
\textit{Avg.} & 
\makecell{\underline{43.28} \\[-1.5pt] \small{(-)}} &
\makecell{\underline{\textbf{43.32}} \\[-1.5pt] \bf \small{(\colorr{+0.04})}} &
\makecell{\underline{43.21} \\[-1.5pt] \small{(-)}} &
\makecell{\underline{\textbf{43.32}} \\[-1.5pt] \bf \small{(\colorr{+0.11})}} \\[-0.5pt] \bottomrule
\end{tabular}
\caption{\label{tab:combination}
The complementary effect of chunk length-based strategies, i.e., ``Chunk (Align.)'' and ``Chunk (LM)'', and monotonicity-based strategy ``+Mcity''. We combine the strategies with significant differences (Covariance$<$0.3) according to correlation analysis in Figure~\ref{fig:corref}: ``+Mcity'' with alignment based chunk length strategy ``Align.'' and language model based chunk length strategy ``LM''.}
\end{table}

\paragraph{SiMT-aware sampling strategies do help.}
We test the effects of our deliberately designed strategies for SiMT. As shown in Table~\ref{tab:sampling_strategies}, we can see that SiMT-aware strategies based on sentence difficulty and chunk length achieve significant improvements against randomly sampling, where the chunk length-based strategies are the most effective (+0.80 and +0.73 BLEU points for ``Align.'' and ``LM'', respectively). Besides, the monotonicity-based strategy ``Mcity'' slightly underperforms the random sampling, especially under high latencies ($k=$5, 7, 9). The potential reason is ``Mcity'' prefers short and word-to-word translations, making the sampled synthetic data intuitively easier. 

To quantitatively investigate the reason for the slightly worse performance for ``Mcity'', we visualize the correlations between ``Mcity'' and other strategies in Figure~\ref{fig:corref}. As shown, the data sampled by the monotonicity-based strategy are significantly different from others. \citet{MonotonicSimultaneousTranslation_2021a} also show that samples chosen by chunk length-based strategy may with poor monotonicity. Given such a huge data gap, it is natural to suspect if there exists a complementary between ``Mcity'' and the best chunk length-based sampling strategies, e.g., chunk length-based strategy.

\paragraph{Chunk length-based and monotonicity-based strategies complement each other.} Based on the above quantitive analysis and suspicion, we combine the chunk length-based strategies and monotonicity-based strategy as follows: 1) sampling monolingual data with the ratio 160\% of the original volume according to the chunk length-based strategy, and 2) reranking the sentences with monotonicity-based strategy, and then filter out the extra 60\%.
As shown in Table~\ref{tab:combination}, we can see that although monotonicity itself does not work well, combining the two gives overall marginal improvements, which is more obvious under low latency, e.g., +0.74 BLEU points improvement on average, indicating the complementary of two types of sampling strategies in difficult scenarios.

Considering the computational complexity of alignment, we will set the LM as the default chunk length-based strategy.
Therefore, we leave the combination of LM-based chunk length strategy and monotonicity-based strategy as the default of our method in the following experiments.

\begin{figure}[t]
\begin{subfigure}{.5\textwidth}
  \centering
  \includegraphics[width=0.7\linewidth]{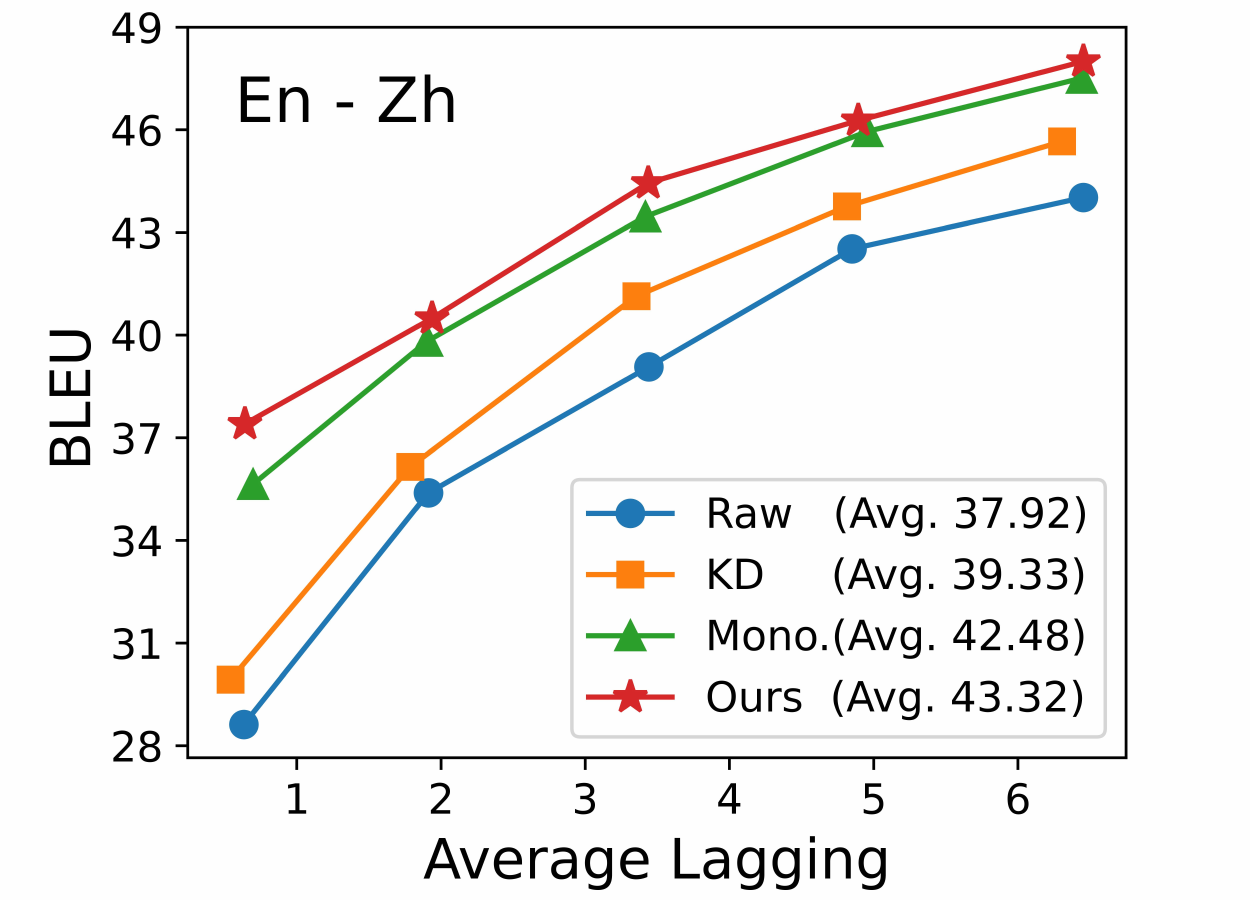}  
\end{subfigure}
\begin{subfigure}{.5\textwidth}
  \vspace{8pt}
  \centering
  \includegraphics[width=0.7\linewidth]{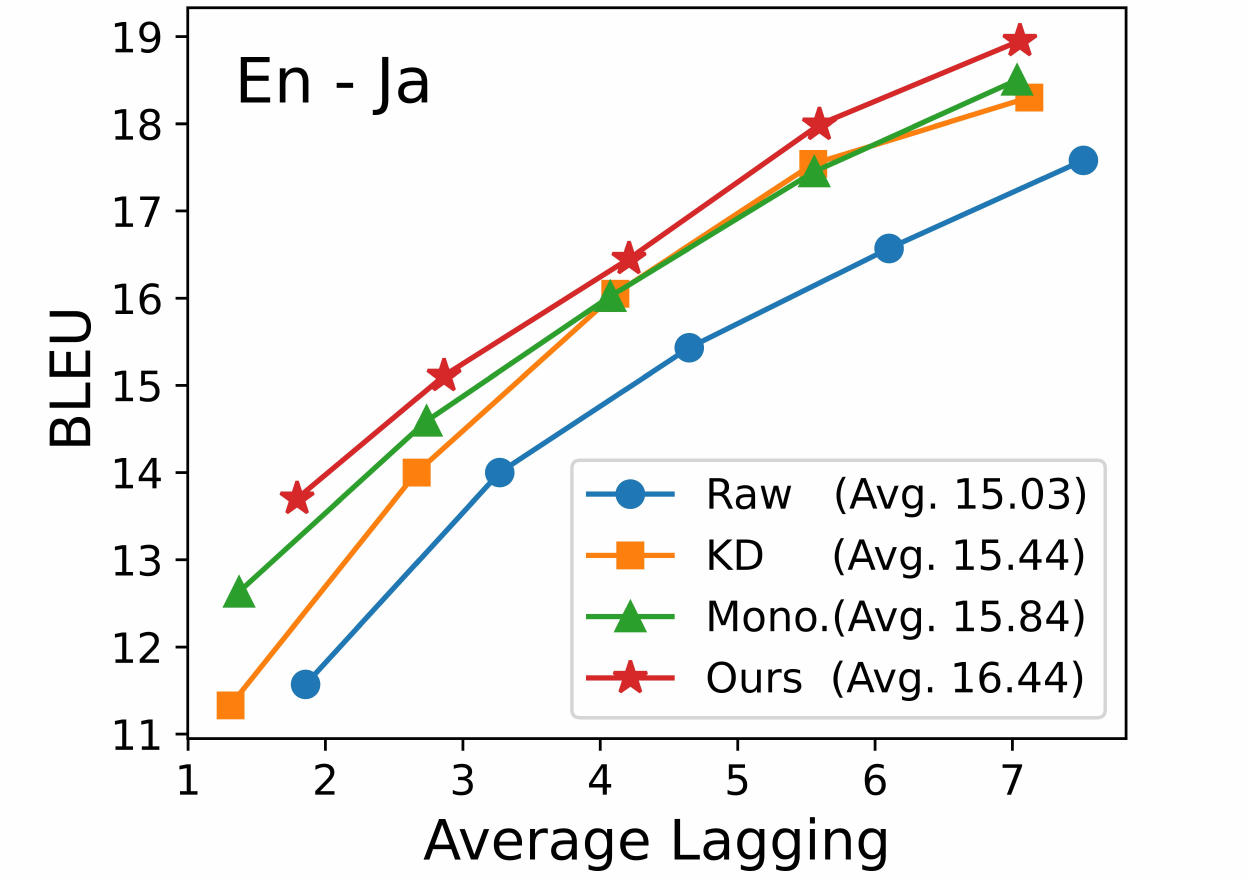}  
\end{subfigure}
\caption{\label{fig:main-results}
Main results on the En$\Rightarrow$Zh (up) and En$\Rightarrow$Ja (down) benchmarks. Each line represents a system, and the 5 nodes correspond to different wait-$k$ settings ($k = 1, 3, 5, 7, 9$). ``Raw'' and ``KD'' represent the systems trained on the original and distilled parallel data, respectively. ``Mono.'' and ``Ours'' demonstrate using monolingual data with the random sampling strategy and our proposed best strategy, respectively.}
\end{figure}

\begin{figure*}[t]
\centering
\includegraphics[width=0.93\textwidth]{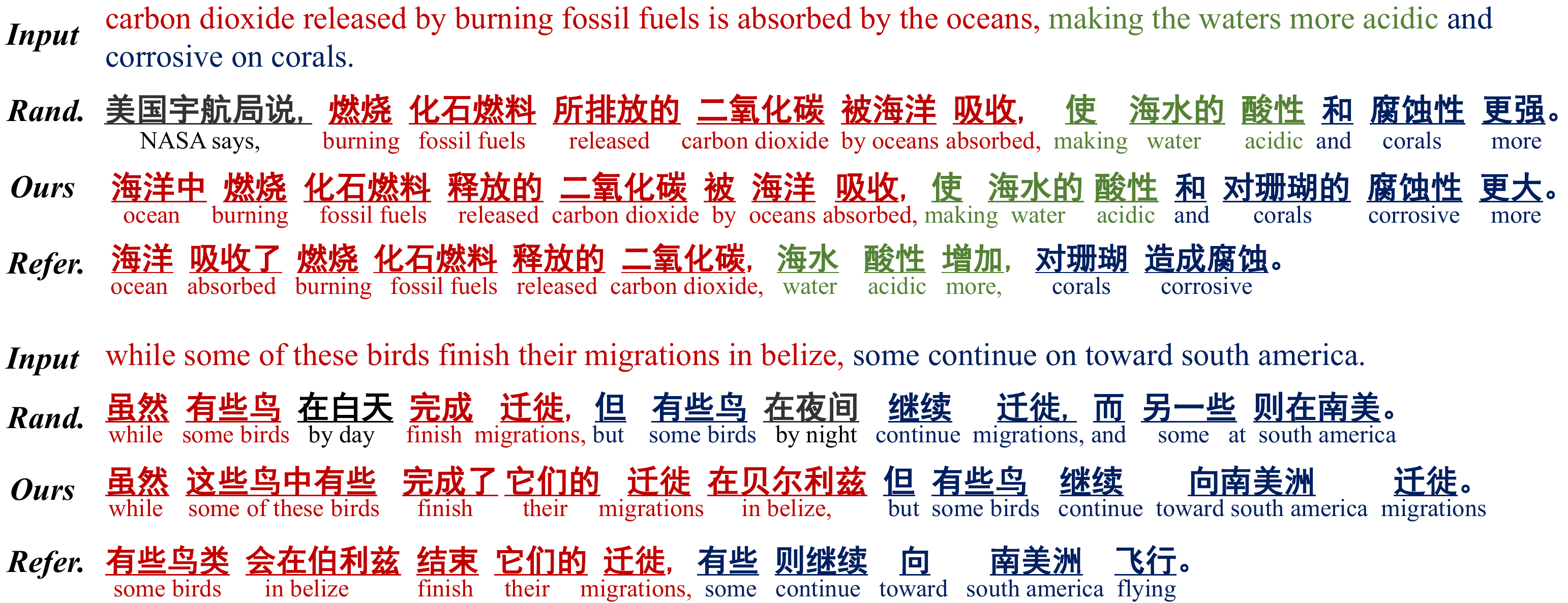}
\caption{\label{fig:example}
Translation examples of models trained with random ``Rand.'' and our ``Ours'' monolingual data sampling strategies under the wait-3 policy. ``Refer.'' means the reference. Words without color are hallucinations.}
\end{figure*}

\subsection{Main Results}

Figure~\ref{fig:main-results} lists the results on the En$\Rightarrow$Zh and En$\Rightarrow$Ja benchmarks, with \textit{average-lagging}~\cite{STACLSimultaneousTranslation_2019} being the latency metric. Encouragingly, the conclusions in the empirical findings hold across language pairs, significantly outperforming the random sampling baseline by +0.84 and +0.60 BLEU points, respectively. This demonstrates the effectiveness and universality of our proposed approach. Notably, our data-level approaches neither modify model structure nor add extra training objectives, thus not changing the latency and maintaining the intrinsic advantages of SiMT models. The main side effect of our approach is the increased inference time for building distilled data with sampled monolingual sentences. Fortunately, the cost is once-for-all, and the distilled synthetic data can be flexibly reused. Given the considerable and consistent SiMT improvement, the above cost is acceptable.

\subsection{Analysis}

In this section, we provide quantitative statistics and qualitative cases to show the superiority of our sampling strategy against random sampling.

\begin{table}[t]
\centering
\begin{tabular}{cccccc}
\toprule
& & \bf\textit{TAnti} & \bf\textit{TCnk} & \bf\textit{GHall} & \bf\textit{GCnk} \\ \midrule
\multirow{2}*{\rotatebox{90}{\small \textbf{EnZh}}} & Rand. & 23.92\small{\%} & 1.11 & 10.69\small{\%} & 1.11 \\
~ & Ours & \bf 13.86\small{\%} & \bf 1.01 & \bf 8.16\small{\%} & \bf 1.08 \\\midrule
\multirow{2}*{\rotatebox{90}{\small \textbf{EnJa}}} & Rand. & 16.47\small{\%} & 1.10 & 6.91\small{\%} & 1.13 \\
~ & Ours & \bf 8.30\small{\%} & \bf 1.02 & \bf 3.08\small{\%} & \bf 1.07 \\ \bottomrule
\end{tabular}
\caption{\label{tab:anapolicy}
{Statistics of monotonicity ``\textit{TAnti}''($\Downarrow$) and chunk length ``\textit{TCnk}''($\Downarrow$) in monolingual training data, and hallucinations ``\textit{GHall}''($\Downarrow$) and chunk length ``\textit{GCnk}''($\Downarrow$) in generations.}}
\end{table}

\begin{table*}[t]
\centering
\vspace{6pt}
\begin{tabular}{lcccccccccc}
\toprule
\bf Scale & \bf Strategy & \bf \textit{wait-1} & \bf\textit{wait-3} & \bf\textit{wait-5} & \bf\textit{wait-7} & \bf\textit{wait-9} & \bf\textit{Avg.} & \textbf{$\mathbf{\Delta}$} & \bf\textit{GHall} & \bf\textit{GCnk} \\ \midrule
\multirow{2}*{1:1} & Rand. & 35.64 & 39.82 & 43.46 & 45.95 & 47.51 & 42.48 & & 10.69\% & 1.11 \\
~ & Ours & \bf 37.40 & \bf 40.49 & \bf 44.44 & \bf 46.27 & \bf 48.00 & \bf 43.32 & \bf +0.84 & \bf 8.16\% & \bf 1.08 \\ \midrule
\multirow{2}*{1:3} & Rand. & 33.79 & 39.26 & 43.48 & \bf 46.27 & \bf 47.84 & 42.13 & & 11.57\% & 1.13 \\
~ & Ours & \bf 36.75 & \bf 41.04 & \bf 44.23 & 45.99 & 47.30 & \bf 43.06 & \bf +0.93 & \bf 7.30\% & \bf 1.09 \\ \midrule
\multirow{2}*{1:5} & Rand. & 35.45 & 39.85 & 43.26 & 46.14 & \bf 47.70 & 42.48 & & 10.79\% & 1.12 \\
~ & Ours & \bf 37.35 & \bf 41.40 & \bf 44.65 & \bf 46.35 & 47.46 & \bf 43.44 & \bf +0.96 & \bf 6.60\% & \bf 1.07 \\ \midrule
\multirow{2}*{1:10} & Rand. & 34.81 & 40.54 & 43.73 & 45.93 & 48.02 & 42.61 & & 10.52\% & 1.12 \\
~ & Ours & \bf 37.33 & \bf 42.25 & \bf 44.00 & \bf 46.62 & \bf 48.09 & \bf 43.66 & \bf +1.05 & \bf 7.26\% & \bf 1.06 \\ \bottomrule
\end{tabular}
\caption{\label{tab:scale-mono}Comparison between random sampling ``Rand.'' and ``Ours'' when scaling up the monolingual data on En$\Rightarrow$Zh. ``Scale'' refers to the proportion of distilled bilingual data and monolingual data. For translation quality, we report BLEU scores (``wait-$k$'' and ``avg.'' $\Uparrow$). For fine-grained evaluation, we report the hallucination rate ``\textit{GHall}'' ($\Uparrow$) and chunk length ``\textit{GCnk}'' ($\Uparrow$) proposed above. We train all models with the same training steps.}
\end{table*}

\begin{table}[t]
\centering
\begin{tabular}{ccccC{1.3cm}}
\toprule
\multicolumn{2}{c}{\bf Teacher} & \bf \textsc{Base}: 48.55 & \bf \textsc{Big}: 51.86 & \textbf{$\mathbf{\Delta}$} \\ \midrule
\multirow{6}*{\rotatebox{90}{\bf Student}} & \textit{wait-1} & 37.40 & \bf 38.22 & \bf +0.82 \\
~ & \textit{wait-3} & 40.49 & \bf 41.84 & \bf +1.35 \\
~ & \textit{wait-5} & 44.44 & \bf 44.65 & \bf +0.21 \\
~ & \textit{wait-7} & 46.27 & \bf 46.35 & \bf +0.08 \\
~ & \textit{wait-9} & 48.00 & \bf 48.34 & \bf +0.34 \\
~ & \textit{Avg.} & 43.32 & \bf 43.88 & \bf +0.56 \\ \bottomrule
\end{tabular}
\caption{\label{better-teacher}
Augmenting the teacher by employing the teacher with a large model capacity (\textsc{Big}) on En$\Rightarrow$Zh.}
\end{table}

Similar to full-sentence NMT, SiMT also suffers from hallucination problem~\cite{HallucinationsNeuralMachine_2018,ImprovingSimultaneousTranslation_2021}, generating fluent but inadequate translations, which is caused by overconfidence of the language modeling~\cite{PreventLanguageModel_2021a}. In SiMT, due to the incomplete source sentence, the contribution of source information in prediction is further reduced, resulting in a more serious hallucination problem~\cite{ImprovingSimultaneousTranslation_2021}. We argue that our strategy is beneficial in avoiding hallucinations, thereby improving the translation quality.

Referring to \citet{ImprovingSimultaneousTranslation_2021}, we use the hallucination rate of hypotheses to evaluate the generation quality, named \textit{GHall}. In more detail, a target word $\hat{y_j}$ is a \textit{hallucination} if it can not be aligned to any source word it can see currently. Formally, based on word alignment $\mathcal{A}$, whether target word $\hat{y_j}$ is a hallucination is:
$$
H(j, \mathcal{A})=\mathds{1}[\{(i, j) \in \mathcal{A} \mid i \geq j + k\}=\varnothing]
$$
The hallucination rate \textit{GHall} is further defined as:
$$
\textrm{GHall}(\boldsymbol{x}, \hat{\boldsymbol{y}}, \mathcal{A})=\frac{1}{|\hat{\boldsymbol{y}}|} \sum_{j=1}^{|\hat{\boldsymbol{y}}|} H(j, \mathcal{A})
$$

We use the same metric as the monotonicity-based strategy to evaluate the monotonicity of the training set averaged over $k \in {1,3,5,7,9}$, named \textit{TAnti}, and the same metric as the chunk length-based strategy based on alignment to evaluate the average length of the training set (\textit{TCnk}) and generations (\textit{GCnk}). 

We first quantitatively analyze how our method affects the constitution of the training data, thereby reducing the translation hallucinations and chunk lengths in Table~\ref{tab:anapolicy}. The anticipation rate and the averaged chunk length of the training data are substantially reduced, leading to a lower hallucination rate and shorter chunks during generation. 
In addition, we give two examples under the wait-3 policy in Figure~\ref{fig:example} to confirm our claim. As seen in the first example, the random sampling strategy generates an unwarranted guess at the speaker ``NASA says,'' and mistranslates the phrase ``on corals'' at the end, while ours perfectly avoids these problems. In the second example, the random sampling strategy mistranslates the phrase ``in belize'', and overconfidencely generates the hallucinated phrases, e.g. ``by day'' and ``by night'', while ours generates adequate and fluent translations.
The above quantitative statistics and qualitative examples demonstrate that our sampling strategy improves the translation against random sampling by reducing the critical issue in SiMT -- hallucination.

\subsection{Scalability Discussion of Our Approach}

In this section, we discuss potential directions to further enhance our scalable method to make SiMT a practical translation system by making the most of the 1) monolingual data, 2) larger teacher, and 3) raw bilingual data. 

\paragraph{Our strategy performs well with more monolingual data.} 
One strength of using monolingual data is the potential to exploit the scaling ability to further improve translation performance~\cite{edunov2018understanding,ding2022redistributing}. To validate our claim, we scale the size of monolingual data by \{$\times$3,$\times$5,$\times$10\} and report the performance of random sampling and ours in Table~\ref{tab:scale-mono}. As seen, enlarging the monolingual data consistently improves the BLEU scores, and with scaling factor increases, our strategy achieves higher performance against random ones, e.g., +1.05 BLEU points under 1:10. Besides, the hallucination rate ``GHall'' and chunk length ``GCnk'' indicate that ours consistently better than that of random sampling, which validates our claim. 

\paragraph{Our strategy performs well with a better teacher.} 
One may expect that augmenting the capacity of the teacher model for our method obtains further improvement. To verify the hypothesis, we employ a larger capacity framework as the teacher, i.e., Transformer-\textsc{Big}.
As shown in Table~\ref{better-teacher}, we see that a larger teacher framework with better translation quality (51.86 vs. 48.55) indeed transfers rich knowledge to the student, further improving the student under all latency settings (+0.56 BLEU points on average).

\begin{table}[t]
\centering
\begin{tabular}{cC{1.7cm}C{1.7cm}C{1.3cm}}
\toprule
 & \bf KD Para. +Mono. & \bf Raw Para. +Mono. & \textbf{$\mathbf{\Delta}$}
 \\[-1pt] \midrule
\textit{wait-1} & 37.40 & \bf 38.13 & \bf +0.73 \\
\textit{wait-3} & 40.49 & \bf 41.72 & \bf +1.23 \\
\textit{wait-5} & \bf 44.44 & 44.38 & -0.06 \\
\textit{wait-7} & 46.27 & \bf 46.61 & \bf +0.34 \\
\textit{wait-9} & \bf 48.00 & 47.82 & -0.18 \\
\textit{Avg.} & 43.32 & \bf 43.73 & \bf +0.41
\\ \bottomrule
\end{tabular}
\caption{\label{tab:rawkdmono}
Replacing the distilled bilingual data (``KD Para.+'') with the raw bilingual data (``Raw Para.+'') in our strategy on En$\Rightarrow$Zh, where ``KD Para.+ Mono.'' is the default setting in the previous experiments.}
\end{table}

\paragraph{Our strategy performs well with raw bilingual data.} Previous experiments in our study make the combination of distilled bilingual data and synthetic data generated by strategically selected monolingual data as default. Although it has shown significantly better performance against the random sampling strategy, all the training data used to train the final SiMT model only utilize the distilled (or synthetic) target-side data, which may lose some long-tailed information in the raw bilingual data~\cite{ding2021rejuvenating,ding2020understanding}. To verify that the raw bilingual data can further complement our monolingual strategy, we replace the distilled bilingual data with the raw one and report the results in Table~\ref{tab:rawkdmono}. We can observe that our strategy performs well with raw bilingual data (+0.41 BLEU points), and the improvements mainly come from the low-latency settings, e.g., +0.73 and +1.23 BLEU points for wait-1 and -3, respectively.

\section{Conclusion}

In this work, we first empirically validate the effectiveness of using monolingual data for SiMT. Then, we propose a simple, effective, and scalable monolingual data sampling strategy, considering both the chunk length and monotonicity. 
Extensive experiments show that our method achieves significant and consistent improvements compared to the random sampling strategy. 
Analyses verify that our strategy improves the translation quality by alleviating the key problems of SiMT, e.g., the hallucination problem. 
Furthermore, our method has appealing expandability and can be further enhanced by 1) enlarging the scale of monolingual data, 2) augmenting the capacity of the teacher, and 3) using the raw bilingual data.

Future directions include 1) validating the effectiveness of our data-level method upon advanced SiMT model~\cite{anonymous2023hidden} and decoding policies~\cite{zhang-etal-2020-learning-adaptive,zhang2022ITpolicy}; and 2) investigating the complementarity~\cite{liu2021complementarity} between our proposed semi-supervised learning based method and the powerful pre-trained models~\cite{liu2020multilingual,zan2022vega} in SiMT.

\section*{Acknowledgments}
We thank the anonymous reviewers for their thorough review and valuable feedback.
Liang and Dacheng were supported by the Major Science and Technology Innovation 2030 ``Brain Science and Brain-like Research'' key project (No. 2021ZD0201405). Xuebo was supported in part by the National Natural Science Foundation of China (Grant No. 62206076 and 62276077) and Shenzhen College Stability Support Plan (Grant No. GXWD20220811173340003 and GXWD20220817123150002).

\bibliography{Mono4SiMT}

\section{Chunk-Extraction Method}

Here, we give a more formal definition and detailed extraction algorithm for the approaches for extracting chunks used in chunk length-based strategy for corpora selection.

For the alignment-based approach, we first match pseudo-targets and compute bidirectional alignments for monolingual data. Then, motivated by phrase-based translation methods in SMT \citep{StatisticalPhraseBasedTranslation_2003,HierarchicalPhraseBasedTranslation_2007}, we extract the shortest contiguously aligned chunk, which is formally definition as follows:

\begin{definition}[Alignment-based chunk]
Given a word-aligned sentence pair $\langle \boldsymbol{x}, \boldsymbol{y}, \sim\rangle$, let $\boldsymbol{x}_{sub}$ be the substring of $\boldsymbol{x}$ with position of its elements make up the set $\mathbb{R}_x$, and the same with $\boldsymbol{y}_{sub}$ and $\mathbb{R}_y$. Then a substring pair $\langle \boldsymbol{x}_{sub}, \boldsymbol{y}_{sub} \rangle$ is a shortest contiguously aligned chunk iff:

1. $\forall i \in \mathbb{R}_x$, $\exists j$, $\boldsymbol{x}_i \sim \boldsymbol{y}_j$, and $\forall \boldsymbol{x}_i \sim \boldsymbol{y}_j$, $j \in \mathbb{R}_y$;

2. $\forall j \in \mathbb{R}_y$, $\exists i$, $\boldsymbol{x}_i \sim \boldsymbol{y}_j$, and $\forall \boldsymbol{x}_i \sim \boldsymbol{y}_j$, $i \in \mathbb{R}_x$;

3. $\boldsymbol{x}_i \nsim \boldsymbol{y}_j$ for all $i \in \mathbb{R}_x$ and $j \notin \mathbb{R}_y$;

4. $\boldsymbol{x}_i \nsim \boldsymbol{y}_j$ for all $i \in \mathbb{R}_x$ and $j \notin \mathbb{R}_y$;

5.$\forall \langle \boldsymbol{x}_{sub}^{\prime} , \boldsymbol{y}_{sub}^{\prime}, \rangle \neq \langle \boldsymbol{x}_{sub}, \boldsymbol{y}_{sub} \rangle$, and $\mathbb{R}_x^{\prime} \subseteq \mathbb{R}_x, \mathbb{R}_y^{\prime} \subseteq \mathbb{R}_y$, it does not meet at least one of the conditions above.
\end{definition}

This definition is close to phrases in SMT, so it can usually extract relatively meaningful chunks. The detailed extraction method is shown in Algorithm~\ref{alg:chunkalign}.

\begin{algorithm}[h]
\caption{Alignment-based method}\label{alg:chunkalign}
\begin{algorithmic}[1]
\Require $\mathcal{A}=\{(i_k,j_k)\}_{k=1}^N$ for all $\boldsymbol{x}_{i_k} \sim \boldsymbol{y}_{j_k}$, and long sentence tendency factor $\alpha$.
\Procedure{calc\_chunk\_Align}{$\mathcal{A},\alpha$}
    \State initialize chunk num $c=0$;
    \While{$\mathcal{A}$ is not Empty}
        \State Sample $(i,j)$ from $\mathcal{A}$
        \State $\mathcal{A},\_$=\Call{remove\_pair}{$\mathcal{A},i,j$}
        \State $c+=1$
    \EndWhile
    \State \Return $S_{chunk}=|\mathcal{A}|^\alpha/c$
\EndProcedure

\Procedure{remove\_pair}{$\mathcal{A},i,j$}
    \State Initialize removed pair list $\mathbb{R}=\{(i,j)\}$
    \State Remove $(i,j)$ from $\mathcal{A}$
    \While{True}
        \State get $i_{max},i_{min},j_{max},j_{min}$ from $\mathbb{R}$
        \For{each pair $(i_k,j_k)$ in $\mathcal{A}$}
            \If{$i_{min} \leq i_k \leq i_{max}$ or $j_{min} \leq j_k \leq j_{max}$}
                \State $\mathcal{A}$,$\mathbb{R}_k$=\Call{remove\_pair}{$\mathcal{A},i_k,j_k$}
                \State $\mathbb{R}=\mathbb{R} \cup \mathbb{R}_k$
                \State Recalculate $i_{max},i_{min},j_{max},j_{min}$
                \State \textbf{break}
            \EndIf
        \EndFor
        \If{$\mathbb{R}$ does not change}
            \State \Return $\mathcal{A}$,$\mathbb{R}$
        \EndIf
    \EndWhile
\EndProcedure
\end{algorithmic}
\end{algorithm}

For the LM-based approach, we consider the continuous part with the highest LM score as a chunk, and when the score drops after adding the next word, it is considered to belong to a new chunk. The detailed extraction method is shown in Algorithm~\ref{alg:chunklm}.

\begin{algorithm}[ht]
\caption{LM-based method}\label{alg:chunklm}
\begin{algorithmic}[1]
\Require monolingual sentence $\boldsymbol{x}$, language model $\phi$, c.
\Procedure{calc\_chunk\_LM}{$\boldsymbol{x},\phi,\alpha$}
\State initialize sentence prefix $\boldsymbol{x}_{sub}=\{\}$;
\State initialize chunk num $c=0$;
\For{each word $x_{i} \in \boldsymbol{x}$}
\State Add $x_{i}$ to $\boldsymbol{x}_{sub}$;
\State $LM^{score}=\phi(\boldsymbol{x}_{sub})$;
\If{$LM^{score}$ decrease}
\State $\boldsymbol{x}_{sub}=\{x_{i}\}$;
\State $c+=1$
\EndIf 
\EndFor 
\State \Return $S_{chunk}=len(\boldsymbol{x})^\alpha/c$
\EndProcedure
\end{algorithmic}
\end{algorithm}

\section{Evaluation with More Translation Metrics}

We report our results under other automatic translation metrics on En$\Rightarrow$Zh, as shown in Table~\ref{other-metrics}. In terms of \textit{BLEURT}~\citep{bleurt} and \textit{chrF2}~\citep{chrf} scores, we can see that our method shows consistently better performance compared to the random sampling baseline, demonstrating the superiority of our approach.

\begin{table}[ht]
\centering
\begin{tabular}{ccccc}
\toprule
\bf Metric & \multicolumn{2}{c}{\bf BLEURT} & \multicolumn{2}{c}{\bf chrF2} \\[-1pt] \cmidrule(lr){1-1}\cmidrule(lr){2-3}\cmidrule(lr){4-5}
\bf Strategy & \bf Rand. & \bf Ours & \bf Rand. & \bf Ours \\ \midrule
\textit{wait-1} & 0.5209 & \bf 0.5325 & 26.07 & \bf 27.11 \\
\textit{wait-3} & 0.5673 & \bf 0.5706 & 28.86 & \bf 29.32 \\
\textit{wait-5} & 0.5954 & \bf 0.5998 & 31.16 & \bf 31.70 \\
\textit{wait-7} & 0.6163 & \bf 0.6197 & 32.91 & \bf 33.02 \\
\textit{wait-9} & 0.6270 & \bf 0.6295 & 33.78 & \bf 34.15 \\
\textit{Avg.} & 
\makecell{\underline{0.5854} \\[-1.5pt] \small{(-)}} &
\makecell{\underline{\textbf{0.5904}} \\[-1.5pt] \bf \small{(\colorr{+0.0050})}} &
\makecell{\underline{30.56} \\[-1.5pt] \small{(-)}} &
\makecell{\underline{\textbf{31.06}} \\[-1.5pt] \bf \small{(\colorr{+0.50})}} \\[-0.5pt] \bottomrule
\end{tabular}
\caption{\label{other-metrics}
\textbf{Main results on the En$\Rightarrow$Zh.} Evaluate under metrics BLEURT and chrF2.}
\end{table}

We also conduct the error-based \textit{MQM}~\citep{DBLP:journals/tacl/FreitagFGRTM21} human evaluation on randomly sampled 100 En$\Rightarrow$Zh translation sentences for the wait-3 policy. The average number of errors in our method is 3.06, while the random sampling baseline is 3.18.

\end{document}